\documentclass[letterpaper, 10 pt, conference]{ieeeconf}  

\IEEEoverridecommandlockouts                              

\usepackage{graphicx}
\usepackage{amsmath} 
\usepackage{amssymb}  
\usepackage{booktabs}
\usepackage{multirow}
\usepackage{cite}
\usepackage{caption}

\title{\LARGE \bf
CA-DGCL: Dynamic Graph Continual Learning via Condensation and Attachment
}

\author{Tingxu Yan$^{1}$ and Ye Yuan*
\thanks{$^{1}$Tingxu Yan and Ye Yuan* are with the College of Computer and
Information Science, Southwest University, Chongqing 400715, China
        {\tt\small a020909@email.swu.edu.cn,yuanyekl@swu.edu.cn}}%
}

\begin{document}

\maketitle
\thispagestyle{empty}
\pagestyle{empty}

\begin{abstract}

Dynamic graph continual learning (DGCL) is an effective manner for handling catastrophic forgetting in dynamic graphs. However, existing DGCL methods underutilize temporal information across graph snapshots. To address this critical issue, we propose a novel framework for Dynamic Graph Continual Learning via Condensation and Attachment (CA-DGCL). Specifically, CA-DGCL first condenses historical graph snapshots into compact semantic representations efficiently. Further, a cross-timestamp node chains is built to construct a third-order tensor and Tucker decomposition is applied to this tensor for obtaining stable node features, which encapsulate historical knowledge. Finally, these node features are used to generate new nodes and attached to the current graph for replaying of past information without compromising the new patterns. In addtion, a refined forgetting measure is introduced to make it more suitable for dynamic graph settings. Extensive experiments demonstrate that CA-DGCL outperforms baselines in forgetting suppression as well as maintain competitive accuracy, proving its efficacy for dynamic graph continual learning.

\end{abstract}

\section{INTRODUCTION}

Graphs serve as fundamental structures for modeling relational data across domains such as social networks. Graph Neural Networks (GNNs) have emerged as the predominant approach for graph representation learning, achieving considerable success. Recent graph-oriented studies have further explored dynamic graph modeling, graph-based clustering, neural community search, graph pooling, and graph-regularized representation learning \cite{b31,b34,b37,b38,b39,b41,b51}. Meanwhile, representative GNN architectures and scalable training strategies, such as attention-based aggregation, expressive graph isomorphism networks, personalized propagation, simplified graph convolution, clustering-based mini-batching, sampling-based training, edge dropping, and self-supervised graph representation learning, have broadened the practical applicability of graph learning \cite{b77,b78,b79,b80,b81,b82,b83,b84}. Nevertheless, real-world graphs are inherently dynamic, continuously evolving through the addition or removal of nodes and edges. This temporal aspect poses significant challenges for conventional GNNs tailored to static graphs, often resulting in catastrophic forgetting where newly learned patterns overwrite previously acquired knowledge.

Although continual learning research has been extended to dynamic graph settings, existing methods do not effectively exploit the temporal information inherent in sequences of graph snapshots to capture intrinsic node characteristics. Related studies on dynamic, spatio-temporal, and tensor-based representation learning demonstrate the importance of modeling temporal evolution and high-order dependencies in non-stationary data \cite{b27,b29,b32,b35,b36,b40,b42,b44,b53,b60,b62}. Dynamic graph representation learning methods further show that evolving node states, temporal self-attention, continuous-time interactions, temporal neighborhoods, and triadic closure are important for capturing non-stationary graph patterns \cite{b85,b86,b87,b88,b89,b90}. Meanwhile, current evaluation frameworks also lack adaptations specific to the properties of dynamic graphs. Metrics borrowed from static domains—such as forgetting rate—measure a model's performance on prior tasks after training on subsequent ones. However, in practical scenarios where tasks are segmented by time in dynamic graphs, retrospective performance on earlier tasks becomes less relevant. Taking node classification as an example, greater emphasis should instead be placed on nodes that persist across multiple tasks.

To address these limitations, we propose the framework for Dynamic Graph Continual Learning via Condensation and Attachment (CA-DGCL). Our approach first condenses historical graph snapshots into compact semantic representations. By constructing tensors from semantic representations across different timestamps and applying tensor decomposition, CA-DGCL uncovers the intrinsic stable features of nodes. Based on these intrinsic stable features, it generates new nodes enriched with historical information, which are then selectively attached to the graph of the current task. This framework is designed to capture the intrinsic features of node communities and preserve historical knowledge in a task-adaptive manner for the current graph structure.

Our contributions are:
\begin{itemize}
\item We propose CA-DGCL, a condensation-attachment framework for dynamic graph continual learning. It condenses historical snapshots into compact semantic representations and attaches new nodes carrying historical information to the current graph, preserving past knowledge while adapting to new patterns.

\item  We shift the evaluation focus from retrospective performance to more practical, persistence-based metrics, better aligning with real-world dynamic graph scenarios. This metric directly reflects whether useful historical knowledge remains effective for nodes that persist across evolving graph snapshots.

\begin{figure*}[!t]  
    \vspace*{\baselineskip} 
	\centering 
	\includegraphics[width=0.8\linewidth]{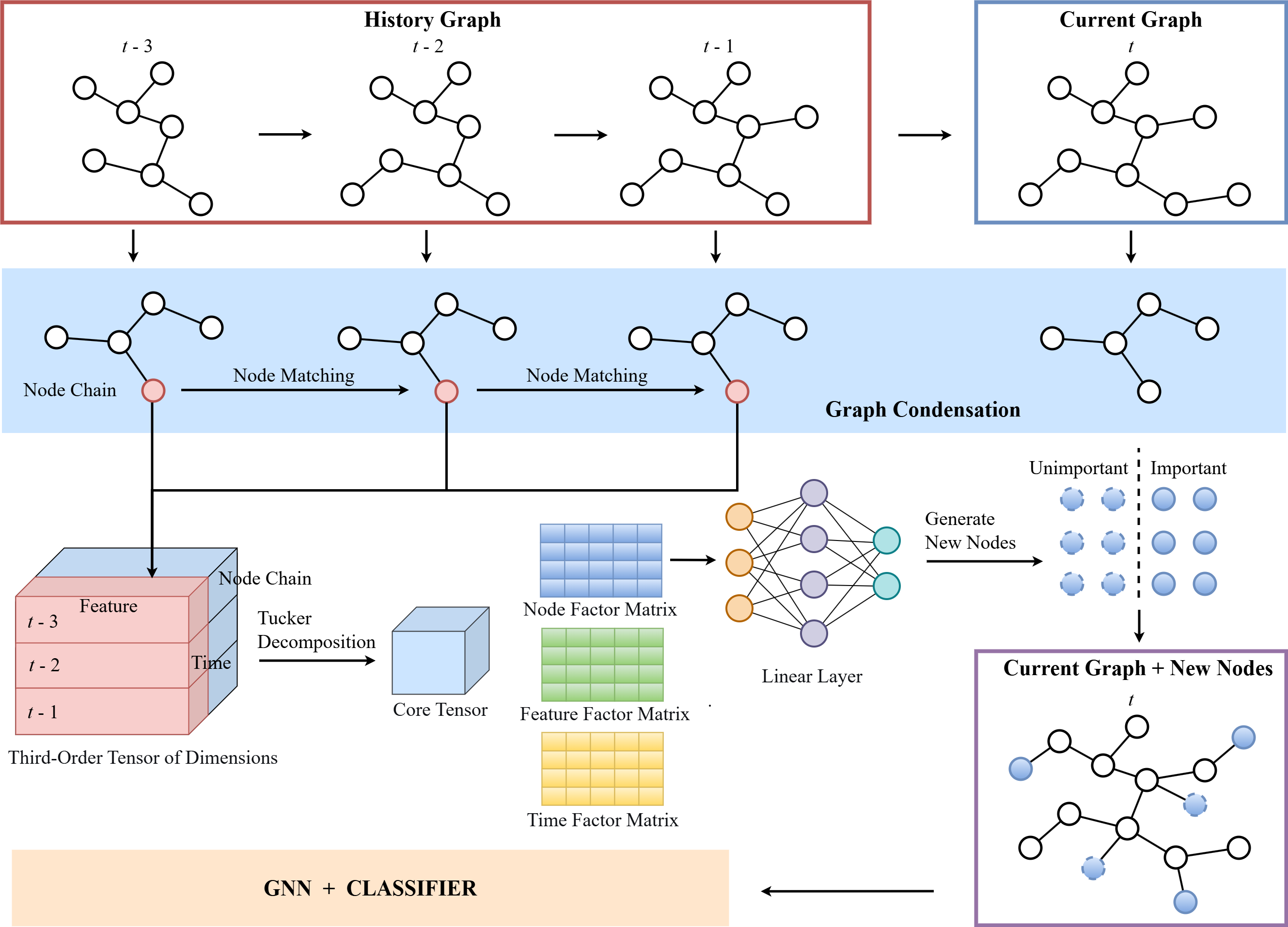}
	\caption{An overview of CA-DGCL. Each task’s graph is condensed to form a sequence of condensed graphs. Through cross-time-step node matching, node chains are constructed, which are then used to build a third-order tensor of dimensions (number of node chains × node feature dimension × time steps). Tucker decomposition is applied to obtain a node factor matrix. This matrix is then mapped back to the original feature space via a linear layer, and the generated new nodes are connected to the current task’s original graph. This linear layer not only performs feature mapping but also learns differentiated mapping weights according to the importance of each node factor to the current task, thereby replaying historical knowledge without affecting the model's plasticity.}  
	\label{fig:example} 
\end{figure*}

\item Extensive experiments on real-world datasets show that CA-DGCL captures long-term node community traits and preserves past knowledge without losing adaptability to new patterns.
\end{itemize}

\section{RLEATED WORK}

\noindent\textbf{Graph Neural Networks.} Graph neural networks (GNNs) \cite{b1} extend deep learning methodologies to data with graph structures, typically by propagating and integrating information from adjacent nodes to generate informative node representations. For instance, the Graph Convolutional Network (GCN) \cite{b2} pioneered an efficient spectral convolution operator based on a first-order approximation of graph Laplacians. Subsequently, GraphSAGE \cite{b3} introduced an inductive framework that relies on sampling and aggregating features from a node's local neighborhood rather than considering all neighbors. In addition, a plethora of other GNN variants \cite{b4,b5,b6,b7,b8,b9,b10,b11,b12,b13} have been proposed, each contributing to substantial progress and achieving strong empirical results across a wide range of graph learning tasks.

\noindent\textbf{Continual Learning.} Continual learning empowers models to assimilate knowledge from a continuous stream of data while safeguarding against the catastrophic forgetting of previous knowledge. Current methodologies generally fall into three primary categories: Regularization-based approaches aim to preserve learned representations by imposing specific constraints on the loss function. For instance, TWP\cite{b14} leverages parameter sensitivity regarding topological structures to apply regularization. Similarly, DyGRAIN\cite{b15} focuses on the receptive field to pinpoint affected nodes. Memory replay-based strategies reinforce knowledge by storing and replaying historical data. Within this domain, ER-GNN\cite{b16} utilizes diverse tactics for selecting nodes to replay. SSM\cite{b17} employs sparsified subgraphs as memory units to maintain topological integrity. DSLR\cite{b18} chooses replay nodes based on coverage metrics and incorporates a link prediction module, while PUMA\cite{b19} condenses original graphs into memory graphs for retraining purposes. Parameter isolation-based techniques dedicate specific parameters to new tasks. PI-GNN\cite{b20} identifies stable parameters via knowledge rectification and expands the network with isolated parameters to capture evolving patterns.

Beyond graph continual learning, general continual learning methods have established important mechanisms for forgetting mitigation, including parameter regularization, distillation, exemplar memory, episodic memory constraints, efficient memory projection, and replay-based experience consolidation \cite{b91,b92,b93,b94,b95,b96}. In addition, latent factor analysis, tensor decomposition, autoencoding, stochastic optimization, and robust representation learning provide useful tools for compactly preserving stable information in high-dimensional data \cite{b28,b30,b33,b43,b45,b47,b48,b49,b50,b52,b54,b55,b56,b58,b59,b61,b63,b64,b65,b66,b67,b68,b69,b70,b71,b72,b73,b74,b75,b76}.

Our proposed framework follows the memory-replay paradigm by extracting
cross-temporal stable features from condensed graphs and
injecting them into the current graph through generated nodes. This design consolidates cross-temporal semantic stability and effectively reduces catastrophic forgetting.

\section{METHOD}
This section presents the details of the CA-DGCL method, as illustrated in Fig.1.

\subsection*{A. Problem Formulation} We define the dynamic graph as a temporal sequence of snapshots, represented by $\mathcal{G}=\{ \mathcal{G}^{(1)}, \mathcal{G}^{(2)},\dots, \mathcal{G}^{(T)} \}$. At any specific time step \textit{t}, the graph snapshot is formulated as $\mathcal{G}^{(t)} = (\mathcal{V}^{(t)}, \mathcal{E}^{(t)})$, where $\mathcal{V}^{(t)} = \{ v_1, v_2, \dots, v_n \}$ denotes the set of nodes and $\mathcal{E}^{(t)} = \{ e_{ij} \mid v_i, v_j \in \mathcal{V}^{(t)} \}$ captures the edges representing their interactions.

Given the non-static nature of the dynamic graph, both the node set $\mathcal{V}^{(t)}$ and edge set $\mathcal{E}^{(t)}$ evolve over time, leading to variations in graph size and composition. To mathematically describe the structural connectivity, we utilize a series of adjacency matrices $\mathcal{A} = \{ \mathcal{A}^{(1)}, \mathcal{A}^{(2)}, \dots, \mathcal{A}^{(t)} \}$. Here, the matrix $\mathcal{A}^{(t)} \in \{0,1\}^{N^{(t)} \times N^{(t)}}$ serves as a binary indicator, where $\mathcal{A}^{(t)}_{ij} = 1$ signifies the existence of an edge $e_{ij}$ within $\mathcal{E}^{(t)}$.

The attributes of the nodes are captured by the feature sequence $\mathcal{X} = \{ \mathcal{X}^{(1)},\mathcal{X}^{(2)}, \dots, \mathcal{X}^{(t)} \}$. The matrix $\mathcal{X}^{(t)} \in \mathbb{R}^{N^{(t)} \times d}$ aggregates the feature vectors for the $N^{(t)}$ nodes present at time \textit{t}, with each node $v_i$ possessing a $d$-dimensional vector $\mathbf{x}_i^{(t)}$. The input fed into the graph neural network at step \textit{t} is the pair $\mathcal{D}^{(t)} = (\mathcal{A}^{(t)}, \mathcal{X}^{(t)})$.
\subsection*{B. Graph Condensation}
Let $\left\{ \mathcal{G}_C^{(1)}, \mathcal{G}_C^{(2)}, \ldots, \mathcal{G}_C^{(T)} \right\}$ represent the sequence of condensed historical graphs, where each $\mathcal{G}_C^{(t)} = (\mathcal{V}_C^{(t)}, \mathcal{E}_C^{(t)}, \mathcal{X}_C^{(t)})$. Our approach adopts the training-free Class-partitioned Graph Condensation framework(CGC)\cite{b21}. CGC first computes the $K$-step propagated features $\mathbf{H} = \hat{\mathbf{A}}^K \mathbf{X}$, where $\hat{\mathbf{A}}$ is the normalized adjacency matrix with self-loops. The condensed node features $\mathbf{X}'$ are then obtained by partitioning each class via an aggregation matrix $\hat{\mathbf{P}} \in \mathbb{R}^{N' \times N}$ (with $N$ and $N'$ being the original and condensed node counts), i.e., $\mathbf{X}' = \hat{\mathbf{P}} \mathbf{H}$. This reduces condensation to a class partition problem, efficiently solvable by clustering without gradient-based optimization. This design is consistent with graph and dataset condensation studies that construct compact synthetic or structure-free representations through gradient matching, trajectory matching, and scalable graph-free condensation \cite{b97,b98,b99,b100}, as well as compact neural representation methods that use tensor compression or low-rank modeling to retain key structural and semantic information \cite{b46,b57}.
\subsection*{C. Generation of New Nodes with Historical Information}
To facilitate the replay of information from condensed historical graphs, we align nodes across temporal snapshots to construct node chains comprising highly similar nodes across different time steps.  By establishing correspondences across adjacent timestamps, we form continuous trajectories that preserve the identity and feature evolution of each node community over time. Subsequently, these node chains are stacked into a third-order tensor with dimensions corresponding to the number of nodes, feature dimension, and time length. The specific methodology is formalized as follows:

\noindent \textbf{Node Chain Construction.} For two consecutive condensed temporal snapshots $\mathcal{G}_{C}^{t-1}$ and $\mathcal{G}_C^t$ with feature matrices $\mathcal{X}_{C}^{t-1} \in \mathbb{R}^{N^{(t-1)} \times d}$ and $\mathcal{X}_C^t \in \mathbb{R}^{N^{(t)} \times d}$, we compute the cosine similarity matrix $\mathbf{S}$ as:
\begin{equation}
\mathbf{S} = \text{norm}(\mathcal{X}_{C}^{t-1})\,\text{norm}(\mathcal{X}_C^t)^\top \in \mathbb{R}^{N^{(t-1)} \times N^{(t)}}
\end{equation}
where $\operatorname{norm}(\cdot)$ denotes $\ell_2$ row-wise normalization. The matching set $\mathcal{M}$ is obtained by:
\begin{equation}
\mathcal{M} = \operatorname{Match}(\mathbf{S}, \tau)
\end{equation}
with $\operatorname{Match}$ being a greedy or Hungarian algorithm that pairs nodes based on the similarity matrix and threshold $\tau$. Matching nodes between condensed graphs at adjacent time steps yields multiple node chains across the sequence.

Based on the matching result, the construction of node chains adheres to a threshold-governed policy. Specifically, if a node pair in $\mathcal{M}$ exhibits a similarity score above the threshold $\tau$, the source node's existing chain from $\mathcal{G}_{C}^{t-1}$ is extended to include the matched target node from $\mathcal{G}_{C}^{t}$. Conversely, any node in $\mathcal{G}_{C}^{t}$ that fails to form a match exceeding $\tau$ is considered a newly emergent entity and thus initiates a new node chain. This policy ensures that chains exclusively represent persistent node communities with consistent feature evolution, while naturally accommodating the birth of novel nodes in the dynamic graph. 

Following the matching process described above, node chains are constructed to track the evolution of individual nodes across the sequence. To maintain consistency, the timeline length of each node chain is always equal to the total number of compressed graphs processed so far. When a new time step arrives, all existing chains first receive a marker indicating absence at the current step. Then, based on the matching set \(\mathcal{M}\) obtained from the current and previous snapshots, the marker of a matched chain is replaced with the actual node identifier from the current compressed graph; unmatched chains retain the absence marker.

Consequently, historical absence markers from previous time steps are never retroactively modified, even if the node reappears later. These past missing entries remain as markers of absence and are later filled during the construction of the evolution tensor. Specifically, for each chain, all observed features are collected. For any time step where the node is absent, the nearest preceding observation is first sought; if none exists, the nearest succeeding observation is used instead. The feature from that nearest observed time step is then employed to fill the missing entry. This design ensures that all chains have uniform length (equal to the number of time steps) while preserving the original record of node appearances and absences, which facilitates modeling the complete temporal evolution via tensor decomposition.

\noindent \textbf{Tensor Construction.} By aligning matched nodes across time steps, we can obtain node chains, and thereby construct a third-order evolution tensor $\boldsymbol{\mathcal{X}} \in \mathbb{R}^{N \times d \times T}$, where $N$ is the total number of node chains, $d$ is the feature dimension, and $T$ is the number of time steps. For each node chain $c$ at time $t$, the feature vector is placed as:
\begin{equation}
\boldsymbol{\mathcal{X}}_{c,:,t} = \mathbf{x}_{c,t} \in \mathbb{R}^d, \quad \forall c = 1,\dots,N,\; t = 1,\dots,T
\end{equation}

\noindent\textbf{Tucker Decomposition.} The tensor $\boldsymbol{\mathcal{X}}$ is decomposed via Tucker decomposition:
\begin{equation}
\boldsymbol{\mathcal{X}} \approx \boldsymbol{\mathcal{G}} \times_1 \mathbf{U} \times_2 \mathbf{V} \times_3 \mathbf{W}
\end{equation}
where $\boldsymbol{\mathcal{G}} \in \mathbb{R}^{R_1 \times R_2 \times R_3}$ is the core tensor, $\mathbf{U} \in \mathbb{R}^{N \times R_1}$ is the node factor matrix, $\mathbf{V} \in \mathbb{R}^{d \times R_2}$ is the feature factor matrix, and $\mathbf{W} \in \mathbb{R}^{T \times R_3}$ is the time factor matrix. The symbol $\times_n$ denotes the $n$-mode product. The decomposition is solved by the Higher-Order Orthogonal Iteration (HOOI) algorithm, which iteratively optimizes:
\begin{equation}
\min_{\boldsymbol{\mathcal{G}},\mathbf{U},\mathbf{V},\mathbf{W}} \big\| \boldsymbol{\mathcal{X}} - \boldsymbol{\mathcal{G}} \times_1 \mathbf{U} \times_2 \mathbf{V} \times_3 \mathbf{W} \big\|_F^2
\end{equation}
subject to $\mathbf{U},\mathbf{V},\mathbf{W}$ having orthogonal columns.

\noindent \textbf{Node Generation and Graph Expansion.} Given the node factor matrix $\mathbf{U} \in \mathbb{R}^{N \times R_1}$ obtained from Tucker decomposition, we map all node factors back to the original feature space via a learnable projection $\mathbf{P} \in \mathbb{R}^{R_1 \times d}$:
\begin{equation}
\mathcal{X}_{\text{new}} = \mathbf{U} \mathbf{P} \in \mathbb{R}^{N \times d}
\end{equation}
These generated nodes are then attached with the original graph via a similarity‑based strategy (e.g., linking each new node to its $k$ most similar original nodes). The resulting extended graph $\mathcal{G}_{\text{ext}} = (\mathcal{V}_{\text{ext}}, \mathcal{E}_{\text{ext}},  \mathcal{X}_{\text{ext}})$ is then fed into the downstream graph neural network encoder. Within each task, we re‑generate ${\mathcal{X}}_{\text{new}}$ and re‑connect the nodes at every training epoch using the current projection $\mathbf{P}$. This encourages the learnable mapping to adaptively emphasize informative node factors and suppress redundant ones, thereby enabling targeted historical replay.

\subsection*{D. Node Embedding}
The expanded graph \(\mathcal{G}_{\text{ext}}^{(t)} = (\mathcal{V}_{\text{ext}}^{(t)}, \mathcal{E}_{\text{ext}}^{(t)}, \mathcal{X}_{\text{ext}}^{(t)})\) is fed into a graph neural network encoder to produce node representations for the current time step. In this work, we adopt GAT as the backbone encoder. Our framework is model-agnostic and can be  extended to other GNN architectures like GCN or GraphSAGE.

Let the node feature matrix of the expanded graph be \(\mathcal{X}_{\text{ext}}^{(t)} \in \mathbb{R}^{N_{\text{ext}} \times d}\), where \(N_{\text{ext}} = N_{\text{orig}} + N_{\text{new}}\) denotes the total number of nodes in the expanded graph, with \(N_{\text{orig}}\) being the number of original nodes and \(N_{\text{new}}\) the number of generated nodes. The corresponding adjacency matrix is \(\mathbf{A}_{\text{ext}}^{(t)} \in \mathbb{R}^{N_{\text{ext}} \times N_{\text{ext}}}\).

GAT updates node representations by leveraging a self-attention mechanism over a node's local neighborhood. Formally, the embedding of node \(v\) at the \(\ell+1\)-th layer is updated as:

\begin{equation}
\mathbf{h}_{v}^{(\ell+1)} = \sigma\!\left( \frac{1}{K} \sum_{k=1}^{K} \sum_{u \in \mathcal{N}(v) \cup \{v\}} \alpha_{vu}^{(k,\ell)} \mathbf{W}^{(k,\ell)} \mathbf{h}_{u}^{(\ell)} \right)
\end{equation}
where \(\mathbf{h}_{v}^{(\ell)}\) is the embedding of node \(v\) at layer \(\ell\), \(\mathcal{N}(v)\) denotes the set of neighboring nodes, \(\mathbf{W}^{(k,\ell)}\) is a learnable weight matrix for the \(k\)-th attention head, \(\alpha_{vu}^{(k,\ell)}\) is the attention coefficient computed via a shared attention mechanism \(\text{LeakyReLU}(\mathbf{a}^{T}[\mathbf{W}\mathbf{h}_v \| \mathbf{W}\mathbf{h}_u])\)---where \(\mathbf{a}\) is a learnable weight vector used to compute the attention score---followed by softmax normalization over neighbors, \(K\) is the number of independent attention heads, \(\|\) denotes concatenation, and \(\sigma(\cdot)\) is a non-linear activation function. After propagating through \(L\) layers, we obtain the final node embeddings for all nodes in the expanded graph:

\begin{equation}
\mathbf{Z}_{\text{ext}}^{(t)} \in \mathbb{R}^{N_{\text{ext}} \times d_{\text{out}}}
\end{equation}
where \(d_{\text{out}}\) is the output embedding dimension.

Since the generated new nodes serve only as carriers of historical knowledge for replay and are not supervised in the current task, we extract the embeddings of the original nodes from the expanded graph:

\begin{equation}
\mathbf{Z}_{\text{orig}}^{(t)} = \mathbf{Z}_{\text{ext}}^{(t)}[1 : N_{\text{orig}}, :] \in \mathbb{R}^{N_{\text{orig}} \times d_{\text{out}}}
\end{equation}

These embeddings are then passed to a classifier:

\begin{equation}
\mathbf{Y} = \mathbf{Z}_{\text{orig}}^{(t)} \mathbf{W}^{(c)}, \quad \mathbf{W}^{(c)} \in \mathbb{R}^{d_{\text{out}} \times |\mathcal{Z}|}
\end{equation}
where \(\mathcal{Z}\) denotes the set of classes. The probability that node \(i\) belongs to class \(z\) is computed as:

\begin{equation}
\hat{y}_{iz} = \frac{\exp(\mathbf{Y}_{iz})}{\sum_{z'=1}^{|\mathcal{Z}|} \exp(\mathbf{Y}_{iz'})}
\end{equation}

Finally, the classification loss is defined as the cross-entropy over the original nodes only:

\begin{equation}
\mathcal{L}_{\text{cls}} = -\frac{1}{N_{\text{orig}}} \sum_{i=1}^{N_{\text{orig}}} \sum_{z=1}^{|\mathcal{Z}|} \mathbf{I}_{iz} \log \hat{y}_{iz}
\end{equation}
where \(\mathbf{I}_{iz}\) is the indicator of the ground-truth class for node \(i\). This design ensures that the model focuses on the nodes that actually exist in the current task during training, while the generated new nodes participate only in message propagation as carriers of historical knowledge, thereby mitigating catastrophic forgetting without introducing additional supervision signals.

\begin{table}[htbp]
    \centering
    \captionsetup{format=plain, labelsep=space, singlelinecheck=true}
    \caption{Statistics of Datasets}
    \label{tab:datasets}
    \begin{tabular}{lrrrr}
        \toprule
        Dataset         & Nodes   & Edges   & Attributes & Tasks \\
        \midrule
        Cora            & 2,708   & 5,429   & 1,433      & 14    \\
        Elliptic        & 203769  & 234355  & 165        & 25    \\    
        DBLP-S          & 20,000  & 75,706  & 128        & 26    \\
        Arxiv-S         & 40,726  & 88,691  & 128        & 13    \\
        \bottomrule
    \end{tabular}
\end{table}

\begin{table*}[!t]
    \vspace*{\baselineskip} 
	\centering
	\captionsetup{format=plain, labelsep=space, singlelinecheck=true}
    \caption{Performance Mean(PM) and Forgetting Measure(FM) of Different Methods for Node Classification}
	\label{tab:4x4_variance}
	\resizebox{\textwidth}{!}{%
		\renewcommand{\arraystretch}{1.5}
		\begin{tabular}{c|cc|cc|cc|cc}
			\hline
			\multirow{2}{*}{\textbf{Method}} & \multicolumn{2}{c|}{\textbf{Arxiv-S}} & \multicolumn{2}{c|}{\textbf{DBLP-S}} & \multicolumn{2}{c|}{\textbf{Cora}} & \multicolumn{2}{c}{\textbf{Elliptic}} \\
			\cline{2-9}
			& PM & FM & PM & FM & PM & FM & PM & FM \\
			\hline
			ER-GNN & $76.63\pm0.00\%$ & $3.48\pm0.00\%$ & $55.41\pm0.00\%$ & $4.35\pm0.02\%$ & $75.75\pm0.89\%$ & $4.93\pm0.39\%$ & $97.13\pm0.06\%$ & $0.41\pm0.00\%$ \\
			ContinualGNN & $34.80\pm2.83\%$ & $32.67\pm5.04\%$ & $50.90\pm0.60\%$ & $8.43\pm1.16\%$ & $77.34\pm2.20\%$ & $3.37\pm0.18\%$ & $93.15\pm0.20\%$ & $0.97\pm0.03\%$ \\
			TWP & $77.31\pm0.10\%$ & $4.04\pm0.17\%$ & $49.29\pm2.23\%$ & $7.20\pm1.09\%$ & $71.59\pm0.50\%$ & $5.75\pm0.72\%$ & $97.35\pm0.05\%$ & $0.10\pm0.01\%$ \\
			PDGNNs-TEM & $72.49\pm0.77\%$ & $6.78\pm0.22\%$ & $55.13\pm0.06\%$ & $5.64\pm0.44\%$ & $71.80\pm1.02\%$ & $5.04\pm0.20\%$ & $\mathbf{97.52\pm0.01\%}$ & $0.42\pm0.01\%$ \\
			Ours & $\mathbf{79.61\pm0.01\%}$ & $\mathbf{3.13\pm0.00\%}$ & $\mathbf{56.02\pm0.08\%}$ & $\mathbf{2.81\pm0.02\%}$ & $\mathbf{77.73\pm0.20\%}$ & $\mathbf{2.23\pm0.06\%}$ & $97.31\pm0.05\%$ & $\mathbf{0.09\pm0.00\%}$ \\
			\hline
		\end{tabular}%
	}
\end{table*}

\begin{figure*}[h]  
	\centering  
    \includegraphics[scale=0.6]{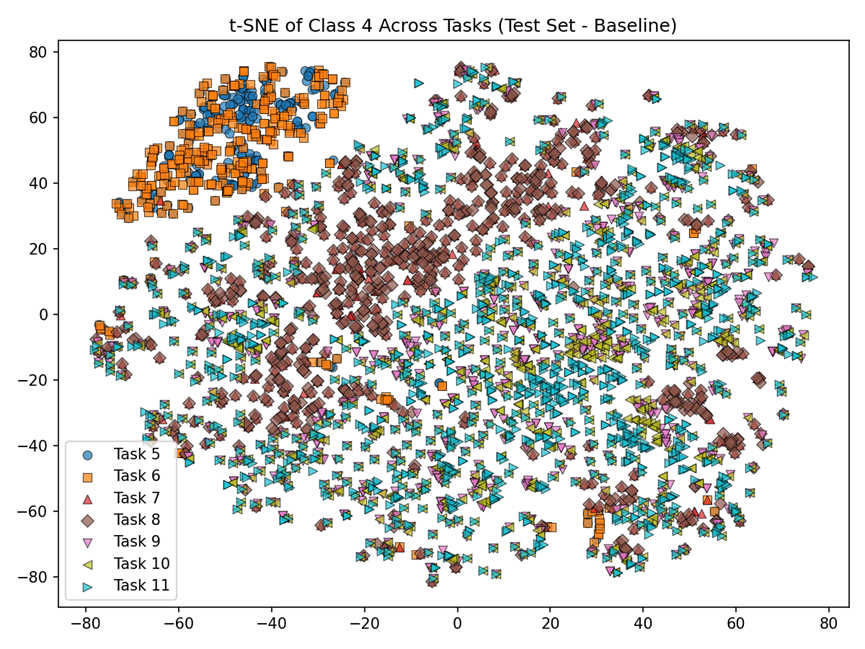}  
    \includegraphics[scale=0.6]{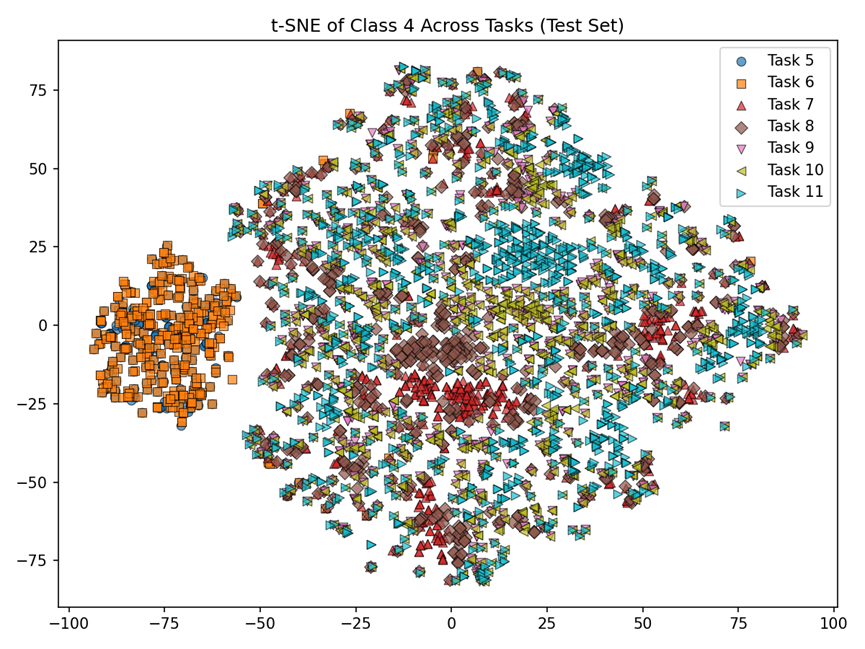}  
	\caption{t-SNE visualization of test node embeddings for a class across different tasks on Arxiv-S. (Left) Fine-tuning (task‑adaptive updating only, no continual learning): both the overall feature space and the intra-task embedding distributions appear relatively loose. (Right) The proposed CA-DGCL model: the overall embedding distribution across tasks remains more compact, and the intra-task clusters are also more concentrated, validating the model's ability to preserve feature space stability.}  
	\label{fig:example}  
\end{figure*}

\section{EXPERIMENTS}

In this section, we present the experimental results of CA-DGCL on four real-world datasets.
\subsection*{A. Setup}
\noindent \textbf{Datasets.} We conduct experiments on three real-world graph datasets with diverse application scenarios: DBLP-S\cite{b22}, Arxiv-S\cite{b20}, Cora\cite{b23} and Elliptic\cite{b24}. Following \cite{b21}, for the streaming graphs DBLP and Arxiv-S, we split them into tasks by timestamps. For the static graphs Cora, we turn it into streaming graphs by adding one new class every two tasks. For the Elliptic, we split it into streaming tasks by its timestamps following our approach. For each dataset, we partition the nodes of $G_T$ into training, validation, and test sets with a 60\%/20\%/20\% split, respectively. This partition is applied consistently across all temporal snapshots. Summary statistics for each dataset are provided in Table~I.

\noindent \textbf{Baselines.} The following baselines are compared:\\
• ContinualGNN\cite{b25} alleviates catastrophic forgetting by detecting new patterns via influence propagation and consolidating existing knowledge through multi-view replay and regularization.\\
• TWP mitigates catastrophic forgetting by preserving the topological structure of the graph and stabilizing important parameters.\\
• ER-GNN  mitigates catastrophic forgetting by storing representative experience nodes from previous tasks and replaying them during the learning of new tasks.\\
• PDGNNs-TEM \cite{b26} addresses the memory explosion problem by decoupling trainable parameters from topological aggregation and compressing computation ego-subnetworks into compact topology-aware embeddings for efficient memory replay.

\noindent \textbf{Metrics.} Two evaluation metrics are adopted: Performance Mean (PM) and Forgetting Measure (FM). PM measures the average model performance:
\begin{equation}PM = \frac{1}{T} \sum_{i=1}^{T} a_i\end{equation}
where \(T\) is the total number of tasks, and \(a_i\) refers to the accuracy of model on task \(i\).

To adapt to dynamic graph scenarios, we modify the traditional forgetting metric. FM measures the ratio of nodes that were correctly predicted in the previous task but incorrectly predicted in the next task. It is defined as the proportion of nodes that are in both the set of correctly predicted nodes in task i and the set of incorrectly predicted nodes in task $i+1$:
\begin{equation}FM = \frac{|C_i \cap E_{i+1}|}{|C_i|}\end{equation}
where $C_i$ represents the set of nodes that were correctly predicted in task \textit{i}, $E_{i+1}$
represents the set of nodes that were incorrectly predicted in task $i+1$ denotes the intersection of the two sets, i.e., the nodes that were correctly predicted in task i but incorrectly predicted in task $i+1$. Traditional forgetting metrics usually work by averaging accuracy drops across all old tasks after training on new ones. However, this method requires keeping data from earlier tasks. This is often not feasible in realistic dynamic graph environments. Furthermore, these retrospective assessments miss the main point of continual learning. The goal is not just to keep performance on obsolete tasks. Instead, it is to sustain useful historical knowledge for current and future inference. Our proposed metric addresses this directly. It measures the continuity of beneficial knowledge retention across adjacent tasks.

\subsection*{B. Overall Results}The experimental results are presented in Table \textrm{II}. In terms of Forgetting Measure (FM), CA-DGCL consistently achieves the lowest values across all four datasets. This demonstrates that the integration of cross‑task node chain modeling with Tucker decomposition of the evolution tensor effectively captures and preserves stable class‑specific semantic patterns, thereby safeguarding previously acquired knowledge against catastrophic forgetting. With respect to Performance Measure (PM), CA-DGCL exhibits highly competitive overall accuracy on every benchmark, ranking either optimal or on par with the strongest baseline. More importantly, this strong discriminative performance is attained in conjunction with the most favorable forgetting suppression—a dual objective that the majority of baselines fail to balance. Typical methods either overfit to the most recent tasks, which degrades FM, or sacrifice plasticity to retain historical knowledge, which compromises PM. CA-DGCL resolves this stability–plasticity trade‑off by extracting intrinsic stable node features from temporal information and leveraging a trainable projection layer to generate new nodes that encode historical knowledge. Critically, this projection layer learns to inject informative historical semantics without perturbing the representation of the current task. Consequently, the augmented graph enriches the model’s contextual awareness while preserving the integrity of the previously learned feature space. This design enables CA-DGCL to simultaneously maintain high accuracy across the task sequence and minimize forgetting, substantiating its effectiveness in continual graph learning scenarios.

\subsection*{C. Cross-Task Stability Of Feature Space}As illustrated in the Fig. 2, the proposed CA-DGCL method demonstrates enhanced cross-task stability in the feature representations of same-class nodes, accompanied by more compact intra-task clustering, when compared to the fine-tuning. These observed improvements indicate that the model has effectively captured the discriminative characteristics of the corresponding node class.

\section{CONCLUSIONS}

In this paper, we proposed CA-DGCL for dynamic graph continual learning.
CA-DGCL condenses historical graphs, constructs cross-temporal node
chains, and applies Tucker decomposition to extract stable node features.
These features are mapped into replay nodes and attached to the current
graph to mitigate catastrophic forgetting. We also introduced a
persistence-based forgetting measure for dynamic graphs. Experiments on
four real-world datasets demonstrate that CA-DGCL consistently reduces
forgetting while maintaining competitive classification performance.

\end{document}